\documentclass[11pt]{article}

\usepackage{fullpage}
\usepackage[ruled, linesnumbered, vlined, commentsnumbered]{algorithm2e}
\usepackage{graphicx,subfig} % figure related
\usepackage{amsfonts,amssymb,amsmath,amsthm,amsopn}	% math related
\usepackage{hhline,booktabs,colortbl,multirow,tabularx,threeparttable} % table related
\usepackage[listings,skins,breakable]{tcolorbox}

\usepackage{enumerate}
\usepackage{authblk}
\usepackage{footnote}
\usepackage{hyperref}
\usepackage{prettyref}
\usepackage{cite}
\usepackage{setspace}
\usepackage{color}
\usepackage{xcolor}  % Required for custom colors
\usepackage{geometry}

\usepackage{tikz}
\usepackage{pgfplots}
\usetikzlibrary{positioning,shapes,shadows,arrows,calc}
\tikzstyle{component}=[rectangle, draw=black, rounded corners, fill=blue!40, drop shadow, text centered, anchor=north, text=white, minimum height=1cm]
\tikzstyle{arrow}=[->, thick]

\pgfplotsset{compat=1.12}
\usetikzlibrary{intersections}
\usetikzlibrary{pgfplots.statistics}
\usepgfplotslibrary{fillbetween}

\def\hlinew#1{%
  \noalign{\ifnum0=`}\fi\hrule \@height #1 \futurelet
   \reserved@a\@xhline}
\makeatother

\definecolor{myblue}{RGB}{34,31,217}
\definecolor{mycyan}{gray}{.7}
\definecolor{Gray}{gray}{0.9}

\DeclareMathOperator*{\argmax}{argmax}
\DeclareMathOperator*{\argmin}{argmin}

\newcommand{\pref}{\prettyref}

\newrefformat{fig}{Fig.~\ref{#1}}
\newrefformat{tab}{Table~\ref{#1}}
\newrefformat{sec}{Section~\ref{#1}}
\newrefformat{app}{Appendix~\ref{#1}}
\newrefformat{alg}{Algorithm~\ref{#1}}
\newrefformat{property}{Property~\ref{#1}}
\newrefformat{theorem}{Theorem~\ref{#1}}
\newrefformat{corollary}{Corollary~\ref{#1}}
\newrefformat{proposition}{Proposition~\ref{#1}}
\newrefformat{def}{Definition~\ref{#1}}
\newrefformat{eq}{equation~(\ref{#1})}

\title{\vspace{-1ex}\LARGE\textbf{An Improved Two-Archive Evolutionary Algorithm for Constrained Multi-Objective Optimization}\footnote{This manuscript is accepted for publication in EMO 2021. The copyright is transferred to Springer.}}

\author[1]{\normalsize Xinyu Shan}
\author[2]{\normalsize Ke Li}
\affil[1]{\normalsize College of Computer Science and Engineering, University of Electronic Science and Technology of China, Chengdu, Sichuan Province, China }
\affil[2]{\normalsize Department of Computer Science, University of Exeter, EX4 4QF, Exeter, UK}
\affil[$\ast$]{\normalsize Email: \texttt{k.li@exeter.ac.uk}}
\date{}

\begin{document}
\maketitle

\vspace{-3ex}
{\normalsize\textbf{Abstract: } Constrained multi-objective optimization problems (CMOPs) are ubiquitous in real-world engineering optimization scenarios. A key issue in constrained multi-objective optimization is to strike a balance among convergence, diversity and feasibility. A recently proposed two-archive evolutionary algorithm for constrained multi-objective optimization (C-TAEA) has be shown as a latest algorithm. However, due to its simple implementation of the collaboration mechanism between its two co-evolving archives, C-TAEA is struggling when solving problems whose \textit{pseudo} Pareto-optimal front, which does not take constraints into consideration, dominates the \textit{feasible} Pareto-optimal front. In this paper, we propose an improved version C-TAEA, dubbed C-TAEA-II, featuring an improved update mechanism of two co-evolving archives and an adaptive mating selection mechanism to promote a better collaboration between co-evolving archives. Empirical results demonstrate the competitiveness of the proposed C-TAEA-II in comparison with five representative constrained evolutionary multi-objective optimization algorithms.

} {\normalsize\textbf{Keywords: } Constrained multi-objective optimization, two-archive evolutionary computation, MOEA/D.}

%!TeX root=main.tex

\renewcommand{\labelitemi}{$\bullet$}

\section{Introduction}
\label{sec:introduction}

The constrained multi-objective optimization problem (CMOP) considered in this paper is formulated as:
\begin{align}\nonumber
    &\mathrm{minimize} \quad \mathbf{F}(\mathbf{x})=(f_1(\mathbf{x}), \cdots, f_m(\mathbf{x}))^T\\
    \begin{split}
     \mathrm{subject\ to}\quad &g_j(\mathbf{x})\geq 0, \quad j=1,\cdots,q\\
    &h_k(\mathbf{x}) = 0, \quad k=1,\cdots,\ell\\
    \end{split}
    \label{eq:cmop}
\end{align}
where $\mathbf{x}=(x_1,\cdots,x_n)^T\in\mathrm{\Omega}$ is a candidate solution. $\mathrm{\Omega}=[a_i,b_i]^n\subseteq \mathbb{R}^n$ as the decision space. $\mathrm{F}:\mathrm{\Omega}\rightarrow \mathbb{R}^m$ consists of $m$ objective functions and $\mathbb{R}^m$ is called the objective space. For a CMOP, the constraint violation of a solution $\mathbf{x}$ can be calculated as:
\begin{equation}
    CV(\mathbf{x})=\sum_{j=1}^{q}\max\{g_j(\mathbf{x}),0\}+\sum_{k=1}^{\ell}\max\{|h_k(\mathbf{x})|-\delta_{k},0\}, 
    \label{eq:cv}
\end{equation}
where $\delta_{k}$ is the tolerance value for the $k$-th equality constraint. $\mathbf{x}$ is regarded as a feasible solution in case $CV(\mathbf{x}=0)$; otherwise it is an infeasible solution. Given two feasible solutions $\mathbf{x}^1$ and $\mathbf{x}^2$ $\in$ $\mathrm{\Omega}$, $\mathbf{x}^1$ is said to Pareto dominate $\mathbf{x}^2$ (denoted as $\mathbf{x}^1 \preceq \mathbf{x}^2$) if $\mathbf{F}(\mathbf{x}^1)$ is not worse than $\mathbf{F}(\mathbf{x}^2)$ at all objectives and $f_i(\mathbf{x}^1) < f_i(\mathbf{x}^2)$ for at least one objective $i \in \{1,\cdots,m\}$. A solution $\mathbf{x}^*$ $\in$ $\mathrm{\Omega}$ is a Pareto-optimal solution of (\ref{eq:cmop}) if and only if there does not exist another solution $\mathbf{x}\in \mathrm{\Omega}$ that dominates $\mathbf{x}^*$. The set of all Pareto-optimal solutions is called the Pareto-optimal set (PS) while their images in the objective space, i.e., $PF=\{\mathbf{F}(\mathbf{x})\in \mathbb{R}^m|\mathbf{x}\in PS\}$, is called the Pareto-optimal front(PF). 

Due to the population-based property, evolutionary algorithms (EAs) have been widely recognized as an effective approach for multi-objective optimization. We have witnessed a significant amount of efforts devoted to the development of evolutionary multi-objective optimization (EMO) algorithms in the past three decades, e.g., fast and elitist multi-objective genetic algorithm (NSGA-II)~\cite{DebAPM02}, indicated-based EA (IBEA)~\cite{ZitzlerK04} and multi-objective EA based on decomposition (MOEA/D)~\cite{ZhangL07}. As stated in our recent paper\cite{LiCFY19}, it is kind of surprising to note that most EMO algorithms are designed for unconstrained MOPs, which rarely exist in real-world engineering optimization scenarios. The lack of adequate research on constraint handling and algorithm design for CMOPs can hinder the wider uptake of EMO in industry. 

It is well known that the balance between convergence and diversity is the cornerstone in EMO algorithm design for unconstrained MOPs. With the consideration of constraints in CMOPs, the feasibility of evolving population becomes an additional issue to handle. As stated in \cite{LiCFY19}, many existing constraint handling techniques in the literature overly emphasize the importance of feasibility thus leading to a loss of diversity and ineffectiveness for problems with many disconnected feasible regions. To strike the balance among convergence, diversity and feasibility simultaneously, Li et al.~\cite{LiCFY19} proposed a two-archive EA (dubbed C-TAEA). Its basic idea is to co-evolving two complementary archives. In particular, one (dubbed as CA) plays as the driving force to push the population towards the PF while the other one (dubbed as DA) mainly tends to maintain the population diversity. Although C-TAEA has shown strong performance on many constrained multi-objective benchmark problems (especially those with many local feasible regions) and a real-world engineering optimization problem. However, partially due to the over emphasis of diversity and the ineffective collaboration between CA and DA, C-TAEA have troubles on problems whose \textit{feasible} PF is dominated by the \textit{pseudo} PF without considering constraints. In this paper, we propose an improved version of C-TAEA (dubbed C-TAEA-II) which maintains the building blocks of C-TAEA but has improved collaboration mechanisms between CA and DA. More specifically, instead of co-evolving both CA and DA separately, the DA in C-TAEA-II is able to progressively guide the CA to escape from local optimal region. In addition, a dynamic mating selection mechanism is proposed to adaptively choose the mating parents from CA and DA according to their evolutionary status. 

The remainder of this paper is organized as follows. \pref{sec:related} provides a pragmatic overview of the current development of EMO for CMOPs along with an analysis of their limitations. The technical details of C-TAEA-II is delineated in \pref{sec:proposal} and its performance is compared with other five state-of-the-art algorithms in \pref{sec:experiments}. \pref{sec:conclusions} concludes this paper and threads some lights of future research.

%!TeX root=main.tex

\vspace{-1.0em}
\section{Background}
\vspace{-0.6em}
\label{sec:related}

In this section, we start with a pragmatic overview of the current important developments of constraint handling techniques in EMO. Thereafter, we discuss some challenges or drawbacks of C-TAEA in two selected CMOPs which constitute the motivation of this paper.

\vspace{-1.0em}
\subsection{Related Works}
\vspace{-0.6em}

In this paper, the existing constraint handling techniques are summarized into five categorizes to facilitate our literature review.
\vspace{-0.6em}
\begin{itemize}
    \item The first group of works derive from the constrained single-objective optimization rigor which mainly rely on penalty functions, e.g., \cite{PandaP16,CoitST96,TessemaY06}. Although this idea is straightforward, its performance is sensitive to the seting of the penalty factor(s) associated with the corresponding penalty function. A too large penalty factor can lead to the indifference of infeasible solutions whereas a too small penalty factor can make infeasible solutions be sanctioned. There have been some attempts to set such penalty factor(s) in an adaptive manner, e.g., \cite{WoldesenbetYT09,HoffmeisterS96}, thus to improve the robustness.
    \item The second category of works mainly rely on feasibility information. Early in \cite{CoelloC99}, Coello Coello and Christiansen proposed a constraint handling method that simply ignores the infeasible solutions. However, this method has no selection pressure when all members of a population are all infeasible ones. Later, Deb et al. proposed a constraint domination principle (CDP) \cite{DebAPM02} which is exactly the same as the classic Pareto dominance when comparing a pair of feasible solutions while infeasbile solutions are compared upon their constraint violations. The CDP becomes one of the most popular constraint handling techniques in EMO afterwards due to its simplicity and effectiveness. For example, Oyama et al. \cite{OyamaSF07} proposed another version of CDP where the dominance relation is compared upon the number of violated constraints. Fan et al. \cite{FanLCHLL16} proposed an angle-based CDP where infeasible solutions within a given angle are compared by using the classic Pareto dominance.
    \item Different from the second category, the third type of approach aims to take advantage of useful infeasible solutions to increase the selection pressure. For example, Peng et al. \cite{PengLG17} proposed evolutionary algorithm based on two sets of weight vectors for CMOPs. In particular, it maintains a set of "infeasible" weight vectors used to select the infeasible solutions with smaller constraint violations. This mechanism helps to preserve a population of well distributed infeasible solutions to facilitate the constrained multi-objective optimization. 
    \item The fourth category is based on $\epsilon$ functions which use a $\epsilon$ as the tolerance threshold for embracing the constraint violations of infeasible solutions (e.g., \cite{TakahamaS12,TakahamaS06}). Note that the setting of $\epsilon$ can largely influence the convergence where a large $\epsilon$ value enforces a large selection pressure towards the convergence while a small $\epsilon$ value implements a fine-tune over a local region.
    \item The last category uses the idea of the interaction between two co-evolving archives to deal with CMOPs. C-TAEA, the baseline algorithm of this paper is a representative example of this category. It maintains two co-evolving archives. One, dubbed CA, works as a regular EMO algorithm while the other one, dubbed DA mainly aims to provide complementary diversity to the CA. The interaction between these two archives is implemented as a restricted mating selection mechanism.
\end{itemize}

\vspace{-1.5em}
\subsection{Drawbacks of C-TAEA}
\vspace{-0.5em}

Although C-TAEA has been reported to have shown a strong performance on a range of constrained benchmark problems, we argue that the interaction or collaboration mechanism between CA and DA is too simple that it only consider the Pareto dominance relation between mating candidates chosen from CA and DA yet ignores their feasibility information. 
\vspace{-1.0em}
\begin{figure}[htbp]
\centering
\includegraphics[width=.7\linewidth]{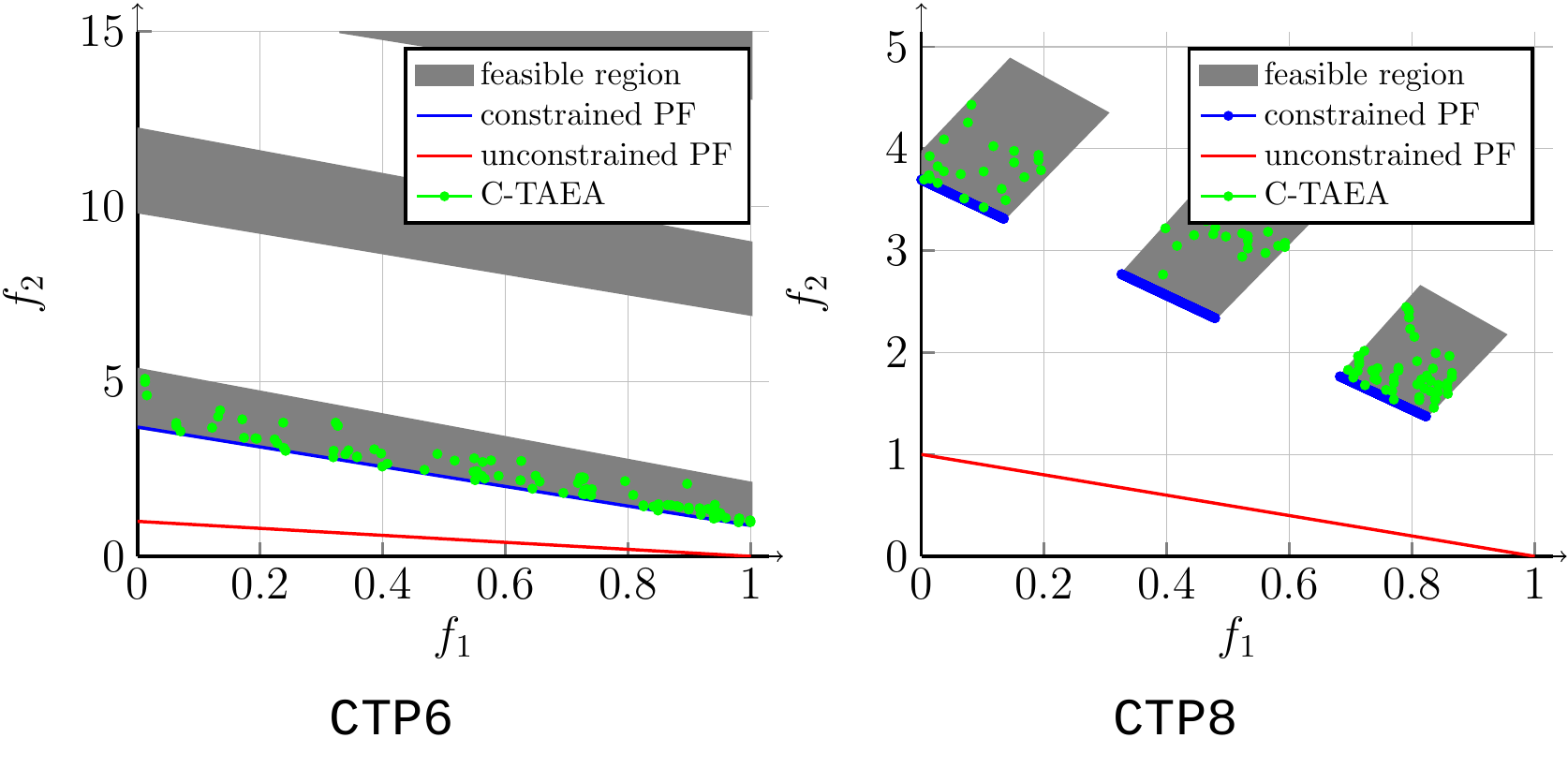}
\caption{Comparative results on the CTP6 and CTP8.}
\label{fig:ctp}
\end{figure}

Let us used two examples shown in \pref{fig:ctp} to further elaborate our above assertion. These are the final populations of solutions obtained by C-TAEA on CTP6 and CTP8 whose \textit{feasible} PF (i.e., the true PF) is dominated by the \textit{pseudo} PF (i.e., the PF without considering constraints) due to the existence of constraints. It is clear that C-TAEA is struggling to converge to the \textit{feasible} PF on both CTP6 and CTP8. In particular, the PF and feasible region of CTP8 is three pieces of disconnected sections while C-TAEA cannot approximate all three sections simultaneously. 

From the above observations, we conclude that if the \textit{feasible} PF is dominated by the \textit{pseudo} PF, mating parents coming from the DA, especially when all members of the DA are infeasible, provide negative guidance on offspring reproduction, thus leading to an ineffective convergence of the CA.

%!TeX root=main.tex

\vspace{-1em}
\section{Proposed Algorithm}
\label{sec:proposal}
\vspace{-0.5em}

\begin{figure}[htbp]
\centering
\includegraphics[width=0.8\linewidth]{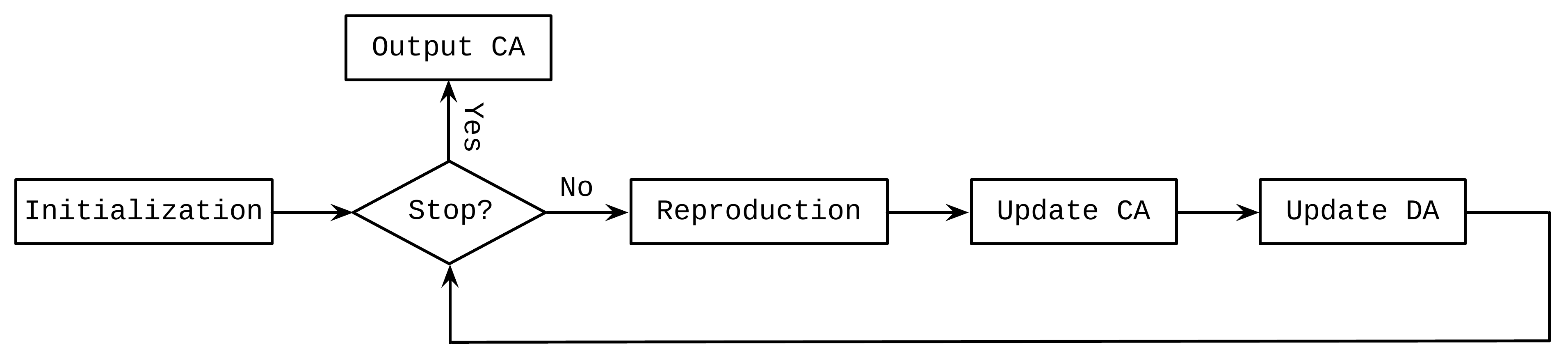}
\caption{Flow chart of C-TAEA-II.}
\label{fig:flowchart}
\end{figure}

The general flow chart of C-TAEA-II, as shown in~\pref{fig:flowchart} follows that of the original C-TAEA. C-TAEA-II maintains two co-evolving archives, dubbed CA and DA, each of which has the same and fixed size $N$. In particular, CA is the driving force that pushes the population to evolve towards the feasible region thus to approximate the \textit{feasible} PF; while DA shows a complementary effect against CA thus to help CA to jump out of the local optima. To leverage the complementary behavior between CA and DA, an adaptive mating selection mechanism is proposed to adaptively select the most suitable mating parents from them according to their evolution status. In the following paragraphs, we describe the update mechanisms of CA and DA as well as the adaptive mating selection mechanism step by step.
\vspace{-1em}
\vspace{-1em}

\vspace{-1em}
\subsection{Update Mechanism of CA}
\label{sec:CA}
\vspace{-0.5em}

CA mainly works at driving the population to evolve towards the \textit{feasible} PF. The pseudo-code of the update mechanism of CA is given in~\pref{alg:updateCA}. Specifically, we first obtain a hybrid population of CA and the offspring population $Q$, dubbed $M_c$ (line 1 of~\pref{alg:updateCA}). In particular, the feasible solutions in $M_c$ is stored in a temporary archive $S_\mathsf{fea}$ while the infeasible ones are stored in another temporary archive $S_\mathsf{infea}$ (lines 2 to 6 of~\pref{alg:updateCA}). Afterwards, there are three different ways to update CA according to the cardinality of $S_\mathsf{fea}$.
\vspace{-0.6em}
\begin{itemize}
    \item If $|S_\mathsf{fea}|$ is lower than $N$, we choose the best $N-|S_\mathsf{fea}|$ infeasible solutions from $|S_\mathsf{infea}|$ to fill the gap according their constraint violations (lines 7 to 10 of~\pref{alg:updateCA}).
    \item If $|S_\mathsf{fea}|$ equals $N$, $|S_\mathsf{fea}|$ is directly used as the next CA (lines 11 and 12 of \pref{alg:updateCA}).
    \item Otherwise, we need to trim $S_\mathsf{fea}$ until its size equals $N$. Similar to C-TAEA, we first initialize a set of evenly distributed weight vectors $W=\{\mathbf{w}^1,\cdots,\mathbf{w}^N\}$, each of which specifies a specific subregion in the objective space $\Delta^i$, $i\in\{1,\cdots,N\}$. To have a well balance between convergence and diversity, each solution in $S_\mathsf{fea}$ is associated with its corresponding subregion as in C-TAEA. Thereafter, the best solution at each subregion having feasible solutions, denoted as $\hat{\mathbf{x}}^i$, $i\in\{1,\cdots,\hat{N}\}$, is chosen to fill the next CA. Specifically, 
    \begin{equation}
        \hat{\mathbf{x}}^i=\argmin\limits_{\mathbf{x}\in\mathrm{\Delta}^i}\{g^\mathsf{tch}(\mathbf{x}|\mathbf{w}^i,\mathbf{z}^{\ast})\},
    \end{equation}
    where
    \begin{equation}
        g^\mathsf{tch}(\mathbf{x}|\mathbf{w}^i,\mathbf{z}^\ast)=\max \limits_{1\leq j\leq m}\{|f_j(\mathbf{x})-z_j^{\ast}|/w_j^i\},
    \end{equation}
    and $\mathbf{z}^\ast$ is the approximated ideal point so far. Note that the number of subregions having feasible solutions $\hat{N}$ equals $N$, these aforementioned selected solutions just fill the new CA. Otherwise, we calculate the absolute difference between the fitness value of the current best solution $\hat{\mathbf{x}}^i$, $i\in\{1,\cdots,\hat{N}\}$, with  respect to the remaining feasible solutions in subregion $\Delta^i$. Afterwards, we iteratively choose the solution having the smallest difference value and use it to fill the new CA. Note that the difference values need to be updated once a solution is selected into the CA.
    
%     Due to there are N subreions, the maximum size of $S$ is N. If the size of $S$ is equal to N, the solutions in $S$ are directly filled into the new CA. Else, we calculate the absolute difference between the fitness value of the best one $\mathbf{x}^n$ of remaining solutions and the selected one $\mathbf{x}^b$ in each subregion. For subregion $\mathrm{\Delta}^i$, They are defined as:
%    \begin{equation}
%        \mathrm{fit}^b=g^{tch}(\mathbf{x}^b|\mathbf{w}^i,\mathbf{z}^{\ast}),
%        \label{eq:fita}
%    \end{equation}
%    where
%    \begin{equation}
%        \mathrm{fit}^n=\min\limits_{\mathbf{x}\in \mathrm{\Delta}^i}\{g^{tch}(\mathbf{x}|\mathbf{w}^i,\mathbf{z}^{\ast})\}.
%    \end{equation}
%     We save the absolute difference of each subregion in $\mathrm{S}^d$. Then, we choose the best one in the subregion which has the smallest difference in $\mathrm{S}^d$ to fill in $\mathrm{S}$. We fill $\mathrm{S}$ in this way until its size is N. This update procedure terminates (lines 26 to 32 of \pref{alg:updateCA}).
\end{itemize}

\begin{algorithm}[htbp]
    \KwIn{$\mathrm{CA}$, offspring population $Q$, weight vector set $\mathrm{W}$} 
    \KwOut{Updated $\mathrm{CA}$}
      
    $S\leftarrow\emptyset$, $S_\mathsf{fea} \leftarrow \emptyset$, $S_\mathsf{infea} \leftarrow \emptyset$;\\
    $M_c \leftarrow \mathrm{CA} \bigcup Q$;\\
    \ForEach{$\mathbf{x}\in M_c$} {
    \uIf{$CV\mathbf{(x)==0}$}
    {
        $S_\mathsf{fea}\leftarrow S_\mathsf{fea}\bigcup \{\mathbf{x}\}$;
    }
    \Else
    {
        $S_\mathsf{infea}\leftarrow S_\mathsf{infea}\bigcup \{\mathbf{x}\}$;
    }
}
\uIf{$|S_\mathsf{fea}|<N$}
{  
    $S\leftarrow S_\mathsf{fea}$;\\
        \While{$|S|<N$}
    {
        $\mathbf{x}^c \leftarrow \argmin\limits_{\mathbf{x}\in S_\mathsf{infea}}\{CV(\mathbf{x})\}$;\\
        $S\leftarrow S \bigcup \{\mathbf{x}^c\}$;
    }
}
\uElseIf{$|S_\mathsf{fea}|==N$}
{
    $S\leftarrow S_\mathsf{fea}$;
}
\Else
{
    %\ForEach{$\mathbf{x} \in S$}
    %{
        %$\overline{\mathbf{F}}_k(\mathbf{x})=\frac{\mathbf{F}(\mathbf{x})-\mathbf{z}^{\ast}}{\mathbf{z}^{nad}-\mathbf{z}^{\ast}}$;
    %}
    Associate each solution in $S_\mathsf{fea}$ to the corresponding subregion $\{\mathrm{\Delta}^1,\cdots,\mathrm{\Delta}^{\hat{N}}\}$ according to the acute angles to predefined weight vectors in $W$;\\
        \For{$i \leftarrow 1$ \KwTo $\hat{N}$}
    {
        $\hat{\mathbf{x}}^i \leftarrow \argmin\limits_{\mathbf{x}\in \mathrm{\Delta}^i}\{g^{tch}(\mathbf{x}|\mathbf{w}^i, \mathbf{z}^{\ast})\}$;\\
        $S\leftarrow S\bigcup \{\hat{\mathbf{x}}^i\}$;\\
        $\Delta^i\leftarrow\Delta^i\setminus\{\hat{\mathbf{x}}^i\}$;\\
    }
    \While{$|S|<N$}
    {
    	\ForEach{$\Delta^i\neq\emptyset, i\in\{1,\cdots,\hat{N}\}$}{
			%Sort the solutions of $\Delta^i$ in ascending order according to their fitness values;\\
			\For{$j\leftarrow 1$ \KwTo $|\Delta^i|$}
			{
				$\delta^i_j\leftarrow |g^{tch}(\hat{\mathbf{x}}^i|\mathbf{w}^i,\mathbf{z}^{\ast})-g^{tch}(\hat{\mathbf{x}}^j|\mathbf{w}^i,\mathbf{z}^{\ast})|$;
			}
			$\delta^i\leftarrow\min\{\delta^i_j\}_{j=1}^{|\Delta^i|}$;\\
		}
		$k\leftarrow\argmin\{\delta^i\}_{i=1}^{\hat{N}}$;\\
		$\bar{\mathbf{x}}\leftarrow\argmin\{\delta^k_j\}_{j=1}^{|\Delta^k|}$;\\
		$S\leftarrow S\bigcup\{\bar{\mathbf{x}}\}$;\\
		$\Delta^k\leftarrow\Delta^k\setminus\{\bar{\mathbf{x}}\}$;\\
    }
}
$\mathrm{CA}\leftarrow S$;\\

\Return $\mathrm{CA}$

\caption{Update Mechanism of CA}
\label{alg:updateCA}
\end{algorithm} 

\vspace{-1.0em}
\subsection{Update Mechanism of DA}
\label{sec:updateDA}

Different from CA, DA does not take the feasibility information into account in its update but mainly aims to improve the diversity of the population as much as possible. The update mechanism of DA is given in~\pref{alg:updateDA}. Like CA, a hybrid population $M_d$ is obtained by combining DA and the offspring population $Q$ (line 1 of~\pref{alg:updateDA}). As in CA, each solution in $M_d$ is associated with its corresponding subregion. Thereafter, the worst solution $\mathbf{x}^w$ from the most crowded subregion $\Delta^c$ is removed from $M_d$ until the size of $M_d$ equals $N$ where
\begin{equation}
    \mathbf{x}^w=\argmax\limits_{\mathbf{x}\in\Delta^c}\{g^\mathsf{tch}(\mathbf{x}|\mathbf{w}^c,\mathbf{z}^{\ast})\}.
\end{equation}

\begin{algorithm}[htbp]
    \KwIn{DA, offspring population $Q$, weight vector set $W$} 
    \KwOut{Updated DA}
    $M_d\leftarrow\mathrm{DA}\bigcup Q$;\\
    Associate each solution in $M_d$ to a subregion forming $\{\mathrm{\Delta}^1,\cdots,\mathrm{\Delta}^N\}$;\\
    \While{$|M_d|>N$}
    {
        Find the most crowded subregion $\mathrm{\Delta}^c$;\\
        $\mathbf{x}^w\leftarrow \argmax\limits_{\mathbf{x}\in \mathrm{\Delta}^c}\{g^\mathsf{tch}(\mathbf{x}|\mathbf{w}^c, \mathbf{z}^{\ast})\}$;\\
        $\mathrm{\Delta}^c \leftarrow \mathrm{\Delta}^c \setminus \{\mathbf{x}^w\}$;\\
        $M_d\leftarrow M_d\setminus\{\mathbf{x}^w\}$;\\
    }
    $\mathrm{DA}\leftarrow M_d$;\\
    \Return $\mathrm{DA}$
\caption{Update Mechanism of DA}
\label{alg:updateDA}
\end{algorithm}

\vspace{-1.5em}
\subsection{Adaptive Mating Selection Mechanism}
\label{sec:mating}

As in C-TAEA, the interaction and collaboration between CA and DA is mainly implemented by the mating selection between them. As discussed in the previous sections, C-TAEA can have trouble with problems whose \textit{pseudo} PF dominates the \textit{feasible} PF. This is mainly attributed to the mating selection mechanism in C-TAEA which is merely built upon the Pareto dominance relationship between solutions chosen from CA and DA. In practice, if the solutions in DA dominate those in CA, solutions coming from DA will be chosen as the mating parents although they may bring negative effect to the evolution process. Even worse, those solutions can potentially lie in the infeasible region thus being useless as mating parents.

In order to address the above problem, we propose an adaptive mating selection mechanism that takes the evolutionary status of CA and DA into consideration. The pseudo-code of this adaptive mating selection mechanism is given in~\pref{alg:mating}. In particular, we maintain a temporary archive $S^t$ to keep a record of the CA from the last generation to track the evolutionary status of CA. More specifically, we compare the Pareto dominance relationship between $S^t$ and the current CA. If the proportion of non-dominated solutions in the current CA (denoted as $\rho_c$) is lower than that of $S^t$ (denoted as $\rho_t$), we set a marker $utility$ as 0; otherwise this marker is set as 1. In the meanwhile, we maintain another marker $choice$, which is initially set as 1. In particular, DA will be used as the mating pool in case $choice$ equals 2; otherwise, i.e., $choice$ equals 1, CA will be chosen as the mating pool. In practice, if $utility$ equals 0, it means that CA may be trapped in a local feasible region. Thus, it is recommended to use DA as the mating pool for offspring reproduction. On the other hand, if DA was used as the mating pool last time and the current $utility$ is still 0, it suggests that DA may go beyond the \textit{feasible} PF thus it is hard to help CA to move any further. In this case, the next mating pool should be chosen as the CA. This alternation between CA and DA as mating pool can help our C-TAEA-II to adaptively choose the most suitable mating parents for offspring reproduction.

\begin{algorithm}[t!]
  \KwIn{CA, DA, $S^t$, $choice$}
  \KwOut{mating parents $p_1$, $p_2$}
  $H_m\leftarrow\mathrm{CA}\bigcup S^t$\;
  $\rho_c \leftarrow$ proportion of non-dominated solution of CA in $H_m$\;
  $\rho_{lc} \leftarrow$ proportion of non-dominated solution of $S^t$ in $H_m$\;
  \uIf{$\rho_{lc} > \rho_c$}
  {
    $utility=0$\;
    \uIf{$choice=1$}
    {
      $choice=2$\;
      $\{p_1,p_2\}\leftarrow$ \texttt{TournamentSelection} $(\mathrm{CA})$\;
%      $\mathrm{p}_2\leftarrow$ \texttt{TournamentSelection} $(\mathrm{DA})$\;
    }
    \Else
    {
      $choice=1$\;
      $\{p_1,p_2\}\leftarrow$ \texttt{TournamentSelection} $(\mathrm{DA})$\;
%      $\mathrm{p}_2\leftarrow$ \texttt{TournamentSelection} $(\mathrm{CA})$\;
    }
  }
  \Else
  {
    $utility=1$\;
  }
  \Return $p_1,p_2$\;
 \caption{Adaptive Mating Selection}
\label{alg:mating}
\end{algorithm}

%!TeX root=main.tex

\vspace{-1.0em}
\section{Empirical Results}
\label{sec:experiments}
\vspace{-1.0em}

To validate the effectiveness of C-TAEA-II, we compare with five representative constrained EMO algorithms, i.e., C-MOEA/D~\cite{JanZ10}, C-TAEA~\cite{LiCFY19}, I-DBEA~\cite{AsafuddoulaRS15}, C-NSGA-III~\cite{JainD14}, C-MOEA/DD~\cite{LiDZK15}. In our experiments, CTP~\cite{DebPM01} and DC-DTLZ~\cite{LiCFY19} are chosen to form the benchmark problem suite. In particular, DC-DTLZ problems are scalable to any number of objectives and are with local feasible regions. Here we use the widely used inverted generational distance (IGD)~\cite{LiCFY19} as the performance metric and Wilcoxon's rank sum test with 5\% significance level is applied to validate the statistical significance of comparison results~\cite{LiZZL09,LiZLZL09,CaoWKL11,LiKWCR12,LiKCLZS12,LiKWTM13,LiK14,CaoKWL14,WuKZLWL15,LiKZD15,LiKD15,LiDZ15,LiDZZ17,WuKJLZ17,WuLKZZ17,LiDY18,ChenLY18,ChenLBY18,LiCFY19,WuLKZZ19,LiCSY19,Li19,GaoNL19,LiXT19,ZouJYZZL19,LiLDMY20,WuLKZ20,BillingsleyLMMG19,LiX0WT20}.

\begin{figure}[htbp]
\centering
    \includegraphics[width=\linewidth]{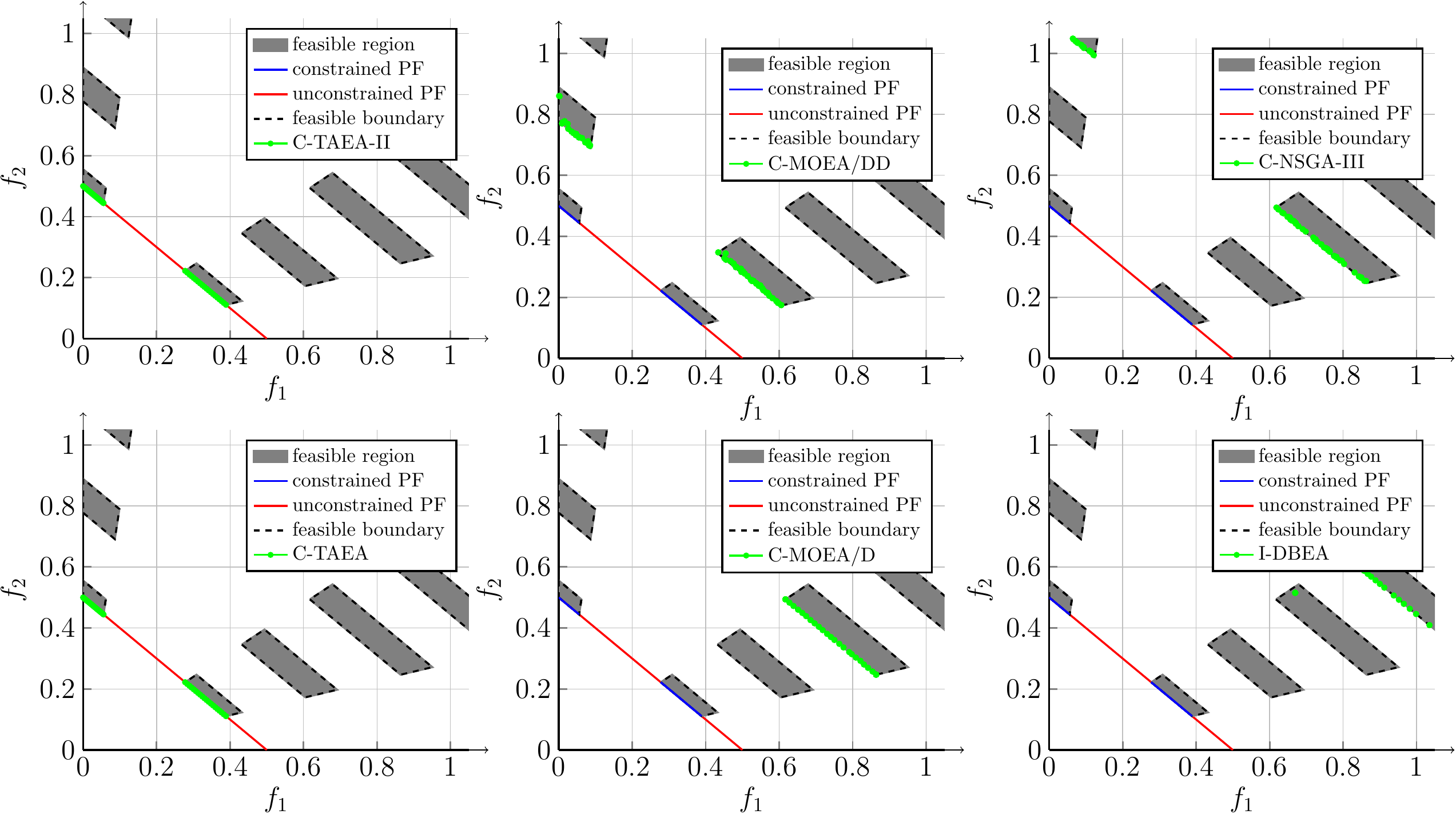}
    \caption{Solutions obtained by C-TAEA-II and the other five peer algorithms with the best IGD metric value on DC3-DTLZ1.}
\label{fig:dc3dtlz1}
\end{figure}

From the comparison results shown in~\pref{tab:ctp} and~\pref{tab:dcdtlz}, we can clearly see the superior performance obtained by C-TAEA-II over the other five peer algorithms including its predecessor C-TAEA. In particular, C-TAEA-II has shown significantly better IGD values on 24 out of 40 ($\sim$ 60\%) test instances on CTP problems. \pref{fig:ctp8}\footnote{Complete results are given in the supplementary document of this paper \url{https://tinyurl.com/yy8jw9bo}} gives an example of solutions obtained by C-TAEA-II and the other five peer algorithms. From this figure, we can see that C-TAEA is struggling to approach the PF while C-TAEA-II can overcome the drawbacks of C-TAEA. As for DC-DTLZ problems, as reported in~\cite{LiCFY19}, these problems are challenging that most existing constrained EMO algorithms, except C-TAEA, are not able to approximate the PF. C-TAEA-II is the best algorithm where it has shown significantly better IGD values on 137 out of 150 ($\sim$ 91.3\%) test instances on DC-DTLZ problems. \pref{fig:dc3dtlz1} gives an example of solutions obtained by C-TAEA-II and the other five peer algorithms. From this figure, we can see that only C-TAEA-II and C-TAEA can find meaningful solutions.

\begin{figure}[htbp]
\centering
\includegraphics[width=\linewidth]{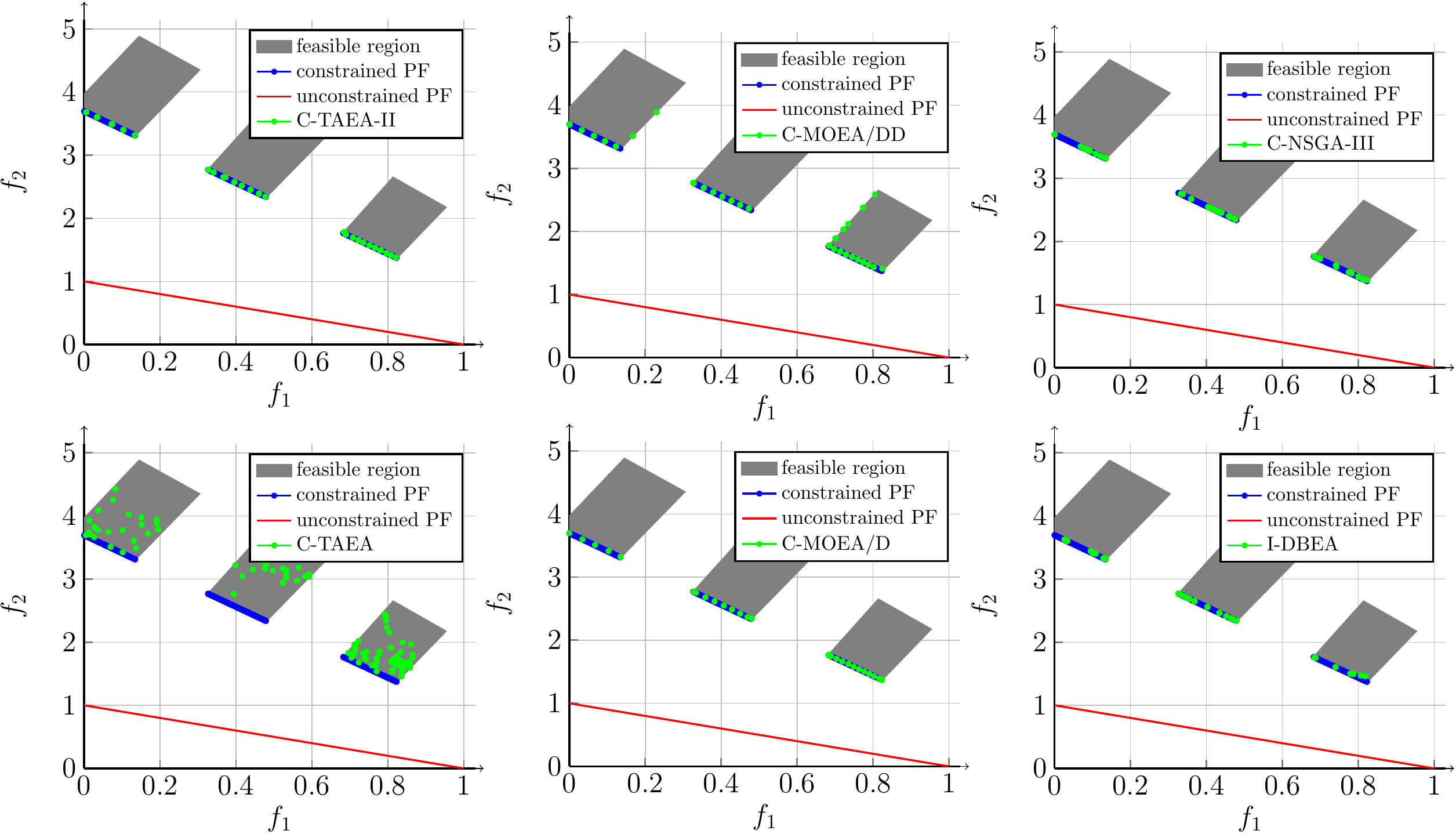}
\caption{Solutions obtained by C-TAEA-II and the other five peer algorithms with the best IGD metric value on CTP8.}
\label{fig:ctp8}
\end{figure}

\begin{table*}[t!]
%    \tiny
    \centering 
    \caption {Comparison results on IGD (median and standard deviation) obtained by C-TAEA-II and the other five peer algorithms on CTP problems.}
    \label{tab:ctp}
    \resizebox{\textwidth}{!}{
    \begin{tabular}{c|c|c|c|c|c|c}
    \hline
         & C-TAEA & C-TAEA-II & C-MOEA/DD & C-MOEA/D & C-NSGA-III & I-DBEA \\ \hline
        CTP1 & 6.39E-3(3.54E-6)$^{\dag}$ & \cellcolor[rgb]{ .651,  .651,  .651} \textbf{4.58E-3(1.52E-8)} & 9.95E-3(3.07E-5)$^{\dag}$ & 4.62E-3(1.45E-9) & 4.06E-2(1.42E-3)$^{\dag}$ & 1.81E-1(3.14E-3)$^{\dag}$ \\ \hline
        CTP2 & 1.23E-2(1.49E-5) & 3.69E-3(2.27E-7) & 6.35E-3(6.79E-7) & \cellcolor[rgb]{ .651,  .651,  .651}\textbf{3.59E-3(2.53E-8)$^{\ddag}$} & 6.95E-3(9.16E-6) & 2.19E-1(1.02E-3) \\ \hline
        CTP3 & 3.39E-2(4.25E-5)$^{\dag}$ & \cellcolor[rgb]{ .651,  .651,  .651}\textbf{7.79E-3(5.43E-5)} & 1.96E-2(8.28E-6)$^{\dag}$ & 1.47E-2(2.01E-10)$^{\dag}$ & 1.14E-2(1.38E-5)$^{\dag}$ & 4.14E-1(1.77E-2)$^{\dag}$ \\ \hline
        CTP4 & 1.87E-1(1.34E-3)$^{\dag}$ & \cellcolor[rgb]{ .651,  .651,  .651}\textbf{5.37E-2(7.94E-5)} & 8.48E-2(1.12E-4)$^{\dag}$ & 5.53E-2(3.37E-6)$^{\dag}$ & 9.23E-2(2.65E-5)$^{\dag}$ & 3.27E-1(2.55E-2)$^{\dag}$ \\ \hline
        CTP5 & 5.84E-3(1.08E-6) & 1.76E-2(8.36E-5) & 2.49E-2(5.52E-6) & 4.72E-3(1.07E-6) & \cellcolor[rgb]{ .651,  .651,  .651}\textbf{4.29E-3(6.49E-7)$^{\ddag}$} & 2.03E-2(7.18E-6) \\ \hline
        CTP6 & 5.47E-2(4.29E-4)$^{\dag}$ & \cellcolor[rgb]{ .651,  .651,  .651}\textbf{1.64E-2(7.21E-8)} & 1.67E-2(7.41E-8) & 1.66E-2(1.41E-9) & 2.99E-2(1.73E-6)$^{\dag}$ & 1.27E-1(1.23E-3)$^{\dag}$ \\ \hline
        CTP7 & \cellcolor[rgb]{ .651,  .651,  .651}\textbf{1.53E-3(2.39E-9)$^{\ddag}$} & 3.32E-3(4.82E-7) & 5.39E-3(1.24E-5) & 3.34E-3(1.18E-10) & 5.68E-3(2.17E-6) & 4.75E-3(6.71E-6) \\ \hline
        CTP8 & 7.03E-2(1.18E-3)$^{\dag}$ & \cellcolor[rgb]{ .651,  .651,  .651}\textbf{1.54E-2(3.84E-6)} & 1.82E-2(1.46E-7)$^{\dag}$ & 1.71E-2(1.59E-7)$^{\dag}$ & 2.66E-2(5.82E-6)$^{\dag}$ & 1.31E-1(2.11E-2)$^{\dag}$ \\ \hline
    \end{tabular}
    }
\begin{tablenotes}
\item[1] $^{\dag}$ denotes the performance of C-TAEA-II is significantly better than the other peers according to the Wilcoxon's rank sum test at a 0.05 significance level; $^{\ddag}$ denotes opposite case.
\end{tablenotes}
\end{table*}

\begin{table*}[t!]
\centering
%\tiny
\caption {Comparison results on IGD (median and standard deviation) obtained by C-TAEA-II and the other five peer algorithms on DC-DTLZ problems.}
\label{tab:dcdtlz}
\resizebox{\textwidth}{!}{
    \begin{tabular}{c|c|c|c|c|c|c|c}
    \hline
        & $m$     & C-TAEA-II & C-TAEA & C-MOEA/DD & C-MOEA/D & C-NSGA-III & I-DBEA \\ \hline
        \multirow{5}{*}{DC1-DTLZ1} & 3     & \cellcolor[rgb]{ .651,  .651,  .651}\textbf{1.51E-2(3.97E-7)} & 5.64E-2(1.96E-6)$^{\dag}$ & 6.09E-2(1.81E-4)$^{\dag}$ & 1.54E-1(2.51E-3)$^{\dag}$ & 6.09E-2(7.41E-6) $^{\dag}$& 5.97E-2(1.99E-4)$^{\dag}$ \\ \cline{2-8}
    & 5     & \cellcolor[rgb]{ .651,  .651,  .651}\textbf{3.72E-2(5.39E-5)} & 6.93E-2(1.85E-4)$^{\dag}$ & 6.46E-2(5.23E-4)$^{\dag}$ & 1.06E-1(1.88E-3)$^{\dag}$ & 8.62E-2(8.36E-4)$^{\dag}$ & 9.97E-2(1.89E-2)$^{\dag}$ \\ \cline{2-8}
    & 8     & \cellcolor[rgb]{ .651,  .651,  .651}\textbf{7.45E-2(4.93E-5)} & 8.46E-2(2.32E-5)$^{\dag}$ & 8.53E-2(3.27E-5)$^{\dag}$ & 8.63E-2(5.37E-3)$^{\dag}$ & 1.07E-1(6.27E-4)$^{\dag}$ & 1.31E-1(5.37E-2)$^{\dag}$ \\ \cline{2-8}
    & 10    & \cellcolor[rgb]{ .651,  .651,  .651}\textbf{7.56E-2(4.43E-5)} & 8.51E-2(5.28E-5)$^{\dag}$ & 8.53E-2(3.71E-5)$^{\dag}$ & 8.62E-2(4.18E-3)$^{\dag}$ & 1.45E-1(4.72E-4)$^{\dag}$ & 1.67E-1(8.42E-2)$^{\dag}$ \\ \cline{2-8}
    & 15    & \cellcolor[rgb]{ .651,  .651,  .651}\textbf{1.13E-1(3.77E-4)} & 1.22E-1(4.92E-4)$^{\dag}$ & 1.18E-1(4.97E-4)$^{\dag}$ & 1.19E-1(6.23E-3)$^{\dag}$ & 1.62E-1(9.24E-3)$^{\dag}$ & 2.32E-1(4.89E-2)$^{\dag}$ \\ \hline\hline
        \multirow{5}{*}{DC1-DTLZ3} & 3     & 4.63E-2(4.97E-5) & \cellcolor[rgb]{ .651,  .651,  .651}\textbf{4.58E-2(1.44E-5)} & 4.97E-2(1.14E-7) & 5.73E-2(4.27E-4) & 5.01E-2(1.98E-5) & 6.82E-2(9.93E-3) \\ \cline{2-8}
    & 5     & \cellcolor[rgb]{ .651,  .651,  .651}\textbf{1.63E-1(8.29E-5)} & 1.65E-1(9.54E-5)$^{\dag}$ & 1.69E-1(6.57E-5)$^{\dag}$ & 6.43E-1(5.63E-4)$^{\dag}$ & 7.26E-1(8.32E-4)$^{\dag}$ & \textbf{\textbackslash{}} \\ \cline{2-8}
    & 8     & \cellcolor[rgb]{ .651,  .651,  .651}\textbf{1.78E-1(1.79E-4)} & 1.81E-1(1.24E-4) & 1.84E-1(3.36E-8)$^{\dag}$ & 6.02E-1(7.83E-3)$^{\dag}$ & 6.99E-1(5.75E-2)$^{\dag}$ & \textbf{\textbackslash{}} \\ \cline{2-8}
    & 10    & 1.92E-1(5.92E-4) & \cellcolor[rgb]{ .651,  .651,  .651}\textbf{1.89E-1(4.72E-4)} & 1.92E-1(7.36E-7) & 6.21E-1(2.92E-5) & 5.85E-1(6.67E-2) & \textbf{\textbackslash{}} \\ \cline{2-8}
    & 15    & \cellcolor[rgb]{ .651,  .651,  .651}\textbf{6.45E-1(2.27E-4)} & 6.51E-1(1.45E-4)$^{\dag}$ & 6.49E-1(5.33E-5)$^{\dag}$ & 8.45E-1(2.51E-3)$^{\dag}$ & 9.69E-1(1.01E-3)$^{\dag}$ & \textbf{\textbackslash{}} \\ \hline\hline
        \multirow{5}{*}{DC2-DTLZ1} & 3     & \cellcolor[rgb]{ .651,  .651,  .651}\textbf{1.86E-2(4.65E-7)} & 1.97E-2(3.56E-7)$^{\dag}$ & \textbackslash{} & \textbackslash{} & \textbackslash{} & \textbackslash{} \\
        \cline{2-8}
    & 5     & \cellcolor[rgb]{ .651,  .651,  .651}\textbf{5.42E-2(3.43E-7)} & 5.44E-2(2.41E-7) & \textbackslash{} & \textbackslash{} & \textbackslash{} & \textbackslash{} \\ \cline{2-8}
    & 8     & 8.94E-2(3.79E-5) & \cellcolor[rgb]{ .651,  .651,  .651}\textbf{8.47E-2(1.23E-5)$^{\ddag}$} & \textbackslash{} & \textbackslash{} & \textbackslash{} & \textbackslash{} \\ \cline{2-8}
    & 10    & \cellcolor[rgb]{ .651,  .651,  .651}\textbf{9.06E-2(3.39E-5)} & 1.04E-1(5.75E-4)$^{\dag}$ & \textbackslash{} & \textbackslash{} & \textbackslash{} & \textbackslash{} \\ \cline{2-8}
    & 15    & \cellcolor[rgb]{ .651,  .651,  .651}\textbf{1.42E-1(3.42E-4)} & 2.18E-1($-$)$^{\dag}$ & \textbackslash{} & \textbackslash{} & \textbackslash{} & \textbackslash{} \\ \hline\hline
        \multirow{5}{*}{DC2-DTLZ3} & 3     & \cellcolor[rgb]{ .651,  .651,  .651}\textbf{6.05E-2(6.47E-6)} & 6.16E-2(8.93E-6)$^{\dag}$ & \textbackslash{} & \textbackslash{} & \textbackslash{} & \textbackslash{} \\
        \cline{2-8}
    & 5     & \cellcolor[rgb]{ .651,  .651,  .651}\textbf{1.57E-1(5.93E-5)} & 1.65E-1(3.96E-5)$^{\dag}$ & \textbackslash{} & \textbackslash{} & \textbackslash{} & \textbf{\textbackslash{}} \\ \cline{2-8}
    & 8     & \cellcolor[rgb]{ .651,  .651,  .651}\textbf{3.21E-1(3.25E-5)} & 1.18E+1($-$)$^{\dag}$ & \textbackslash{} & \textbackslash{} & \textbackslash{} & \textbf{\textbackslash{}} \\ \cline{2-8}
    & 10    & \cellcolor[rgb]{ .651,  .651,  .651}\textbf{3.41E-1(3.92E-5)} & 3.87E-1($-$)$^{\dag}$ & \textbackslash{} & \textbackslash{} & \textbackslash{} & \textbf{\textbackslash{}} \\ \cline{2-8}
    & 15    & \cellcolor[rgb]{ .651,  .651,  .651}\textbf{7.29E-1(2.39E-3)} & 7.93E-1($-$)$^{\dag}$ & \textbackslash{} & \textbackslash{} & \textbackslash{} & \textbf{\textbackslash{}} \\ \hline\hline
        \multirow{5}{*}{DC3-DTLZ1} & 3     & \cellcolor[rgb]{ .651,  .651,  .651}\textbf{2.27E-2(3.73E-6)} & 4.47E-2(1.23E-5)$^{\dag}$ & 3.35E-1(3.89E-2)$^{\dag}$ & 2.75E+0(5.09E+0)$^{\dag}$ & 5.01E-1(1.26E-1)$^{\dag}$ & 6.19E-1(2.65E-1)$^{\dag}$ \\ \cline{2-8}
    & 5     & \cellcolor[rgb]{ .651,  .651,  .651}\textbf{6.31E-2(7.75E-5)} & 7.89E-2(2.58E-3)$^{\dag}$ & 3.91E-1(2.01E-2)$^{\dag}$ & 3.18E-1(5.71E-2)$^{\dag}$ & 2.22E-1(8.39E-2)$^{\dag}$ & 4.92E-1(3.97E-1)$^{\dag}$ \\ \cline{2-8}
    & 8     & \cellcolor[rgb]{ .651,  .651,  .651}\textbf{8.63E-2(4.86E-5)} & 1.81E-1(3.49E-3)$^{\dag}$ & 5.89E-1(3.97E-2)$^{\dag}$ & 5.73E-1(6.83E-1)$^{\dag}$ & 5.62E-1(9.43E-2)$^{\dag}$ & 6.82E-1(6.87E-1)$^{\dag}$ \\ \cline{2-8}
    & 10    & \cellcolor[rgb]{ .651,  .651,  .651}\textbf{1.65E-1(5.32E-5)} & 2.07E-1(3.74E-4)$^{\dag}$ & 7.78E-1(4.93E-2)$^{\dag}$ & 7.93E-1(9.87E-1)$^{\dag}$ & 9.03E-1(6.48E-1)$^{\dag}$ & 9.74E-1(8.36E-1)$^{\dag}$ \\ \cline{2-8}
    & 15    & \cellcolor[rgb]{ .651,  .651,  .651}\textbf{1.81E-1(5.38E-5)} & 5.23E-1(4.83E-4)$^{\dag}$ & 1.25E+0(7.49E-2)$^{\dag}$ & 1.08E+0(8.43E-1)$^{\dag}$ & 1.37E+0(9.34E-1)$^{\dag}$ & 1.27E+0(6.38E+0)$^{\dag}$ \\ \hline\hline
        \multirow{5}{*}{DC3-DTLZ3} & 3     & \cellcolor[rgb]{ .651,  .651,  .651}\textbf{7.23E-2(4.68E-5)} & 1.12E-1(1.31E-4)$^{\dag}$ & 1.19E+0(2.03E+0)$^{\dag}$ & 1.05E+1(8.49E+1)$^{\dag}$ & 1.45E+0(1.05E+0)$^{\dag}$ & 1.15E+0(9.57E-1)$^{\dag}$ \\ \cline{2-8}
    & 5     & \cellcolor[rgb]{ .651,  .651,  .651}\textbf{9.17E-2(3.95E-5)} & 1.77E-1(2.36E-4)$^{\dag}$ & 2.83E-1(1.14E-2)$^{\dag}$ & 2.66E+0(4.57E+0)$^{\dag}$ & 1.34E+0(1.62E+0)$^{\dag}$ & \textbf{\textbackslash{}} \\ \cline{2-8}
    & 8     & 2.51E-1(4.79E-5) & \cellcolor[rgb]{ .651,  .651,  .651}\textbf{2.24E-1(6.54E-5)$^{\ddag}$} & 4.32E-1(2.31E-2) & 2.65E+0(1.18E+1) & 2.09E+0(1.85E+0) & \textbf{\textbackslash{}} \\ \cline{2-8}
    & 10    & \cellcolor[rgb]{ .651,  .651,  .651}\textbf{2.57E-1(5.47E-5)} & 2.75E-1(5.86E-5)$^{\dag}$ & 2.76E-1(1.36E-2)$^{\dag}$ & 7.82E-1(6.23E-2)$^{\dag}$ & 2.06E+0(2.11E+0)$^{\dag}$ & \textbf{\textbackslash{}} \\ \cline{2-8}
    & 15    & \cellcolor[rgb]{ .651,  .651,  .651}\textbf{6.76E-1(7.49E-5)} & 6.85E-1(7.28E-4)$^{\dag}$ & 1.16E+0(7.09E-2)$^{\dag}$ & 1.14E+0(4.49E-2)$^{\dag}$ & 2.11E+0(4.21E+0)$^{\dag}$ & \textbf{\textbackslash{}} \\ \hline
        \end{tabular}%
}
\begin{tablenotes}
\item[1] $^{\dag}$ denotes the performance of C-TAEA-II is significantly better than the other peers according to the Wilcoxon's rank sum test at a 0.05 significance level; $^{\ddag}$ denotes the corresponding algorithm significantly outperforms C-TAEA-II. $\setminus$ denotes the corresponding fails to find meaningful solutions, while $-$ denotes the standard deviation is not available.

\end{tablenotes}    
\end{table*}

%!TeX root=main.tex

\section{Conclusions and Future Works}
\label{sec:conclusions}

In this paper, we proposed an improve version of C-TAEA, a state-of-the-art constrained EMO algorithm. Our proposed C-TAEA-II derives from the basic flow of C-TAEA but is featured with two distinctive modifications. The update mechanisms of CA and DA are improved in order to amplify the complementary behavior of both archives. In addition, an adaptive mating selection mechanism is proposed to avoid choosing infeasible solutions beyond the \textit{feasible} PF. From our preliminary experiments on CTP and DC-DTLZ problems, we observed encouraging results that C-TAEA-II has shown statistically better IGD values on 85\% comparisons. As for the next step, we plan to further improve the collaboration mechanism between two co-evolving archives and we aim to conduct comprehensive validation on a wider range of benchmark problems and real-world engineering optimization problems.

%In this paper, we propose an enhanced two-archive constrained optimization algorithm (C-TEST). In C-TEST, the two populations adopted different evolutionary mechanisms. The constrained-oriented archive is defined as CA, the unconstrained-oriented archive is defined as UA. At each iteration, the parent population is changed when the convergence utility of CA is zero. When CA convergence is effective, the parent population does not change. In this way, we can find out whether the CA falls into local optimal in time. At the same time, we can determine the stable convergence when CA has entered the global optimal region. The performance of C-TEST has been investigated on a series of benchmark problems. We compared with five state-of-the-art CMOAs. The CMOPs considered in this paper do not embrace all types of constraints in the real-world. We hope that our work will lead more people pay attention to the interaction between the two archives.     

\section*{Acknowledgement}
K. Li was supported by UKRI Future Leaders Fellowship (Grant No. MR/S017062/1).

\bibliographystyle{IEEEtran}
\bibliography{IEEEabrv,cmo}

\end{document}